\documentclass{ecai}
\usepackage{graphicx}
\usepackage{xspace}
\usepackage{algorithm}
\usepackage[noend]{algorithmic}
\usepackage{amsmath}
\usepackage{float}

\ecaisubmission

\begin{document}
\title{Disturbing Reinforcement Learning Agents \\ with Corrupted Rewards}
\author{Rubén Majadas \and Javier García  \and Fernando Fernández\institute{This paper has been accepted in RAISA3 workshop celebrated in ECAI 2020 conference. Departamento de Informática, Universidad Carlos III de Madrid Avda. de la Universidad, 30. 28911 Leganés (Madrid). Spain Email addresses: rmajadas@pa.uc3m.es (Rubén Majadas), fjgpolo@inf.uc3m.es (Javier García) and ffernand@inf.uc3m.es (Fernando Fernández)}}

\maketitle              
\begin{abstract}
Reinforcement Learning (RL) algorithms have led to recent successes in solving complex games, such as Atari or Starcraft, and to a huge impact in real-world applications, such as cybersecurity or autonomous driving. In the side of the drawbacks, recent works have shown how the performance of RL algorithms decreases under the influence of soft changes in the reward function. However, little work has been done about how sensitive these disturbances are depending on the aggressiveness of the attack and the learning exploration strategy. In this paper, we propose to fill this gap in the literature analyzing the effects of different attack strategies based on reward perturbations, and studying the effect in the learner depending on its exploration strategy. In order to explain all the behaviors, we choose a sub-class of MDPs: episodic, stochastic \textit{goal-only-rewards} MDPs, and in particular, an intelligible grid domain as a benchmark. In this domain, we demonstrate that smoothly crafting adversarial rewards are able to mislead the learner, and that using low exploration probability values, the policy learned is more robust to corrupt rewards. Finally, in the proposed learning scenario, a counterintuitive result arises: attacking at each learning episode is the lowest cost attack strategy.

\end{abstract}
\section{Introduction}

Reinforcement Learning (RL) is a type of machine learning whose main goal is finding an action policy that makes the agent act optimally in an environment, generally formulated as a Markov Decision Process (MDP). Recently, RL has been successfully used in complex computer games (e.g., Atari~\cite{mnih2013playing}, or Starcraft~\cite{alphastarblog2}), but also in real-world applications where robustness is a key factor (e.g., robotic tasks~\cite{DBLP:journals/jmlr/LevineFDA16}, cyber security~\cite{ferdowsi2018robust} or autonomous driving~\cite{kuderer2015learning}). However, the robustness of RL algorithms is in doubt due to recent works showing that they are vulnerable to \textit{adversarial attacks}. In Adversarial RL~\cite{tong2019}, the most usual of these attacks tries to slightly change the observation (i.e., the state) that the agent perceives from the environment~\cite{behzadan2017vulnerability}, similarly to adversarial examples for classifiers~\cite{DBLP:journals/corr/SzegedyZSBEGF13}. Such attacks can be carried out during the training~\cite{kos2017delving} or the test phase~\cite{DBLP:journals/corr/HuangPGDA17}, and the objective is to disturb as little as possible the original state so as not to be detected, while achieving a system failure, i.e., to make the agent believe that it is in a situation when it is really in another one.

Besides the state perception, another sensitive target to attack is the reward function. In real-world applications, we can obtain rewards that differ from the original due to several factors such as incorrect measurements from sensors, noise in the transmission channel, but also due to attacks that intentionally perturb the reward signal.  However, this attack source has received very little attention~\cite{romoff2018reward,DBLP:journals/corr/abs-1810-01032}, and as far as we know, there are no studies about how sensitive the learning process is depending on the aggressiveness of the reward perturbations and the exploration strategy. In spite of that, we consider that such an analysis is essential for everyone who works in the emerging area of Adversarial RL both designing attacks and defense mechanisms. 

In this paper, we contribute to fill this gap in the literature. In particular, we address a sub-class of MDPs: episodic, stochastic \textit{goal-only-rewards} MDPs~\cite{reinke2017}. For \textit{goal-only-reward} MDPs, a reward is only given if a goal state is reached. Although this is a strict restriction, many decision tasks are modeled with \textit{goal-only-rewards} from classical control problems such as the \textit{Mountain Car} or the \textit{Cart Pole}, to more complex problems such as real robot manipulation tasks~\cite{morere18a}. However, we focus on an intelligible 20$\times$20 grid domain with goal states and obstacles. Focusing on an on-policy RL algorithm, SARSA, and two well-known exploration strategies ($\epsilon$-greedy  and softmax), this paper address a primary question: \textit{how the disturbances in the reward function when a goal state is reached affect the convergence of a learning agent depending on the exploration strategy selected?}

Therefore, our contribution is twofold: first, a threat model for the reward function in \textit{goal-only-rewards} MDPs; second, a comprehensive study in order to analyze the robustness of the learning process depending on the probability of reward perturbations and the exploration strategy. Results show that small exploration probabilities are more resilient to adversary presence, and that attacking at each learning episode is the lowest cost attack strategy.

\begin{section}{Background}

In the literature, we identify three different \textit{attack targets} where one can inject Adversarial Examples in any RL algorithm. 
Figure~\ref{fig:RLproccess} shows these attack targets where each adversary represents a different kind of attack. These possible attack points are the state perceived, the action selected or the reward function:
\begin{figure}[htb!]
    \centering
    \includegraphics[width=0.9\columnwidth]{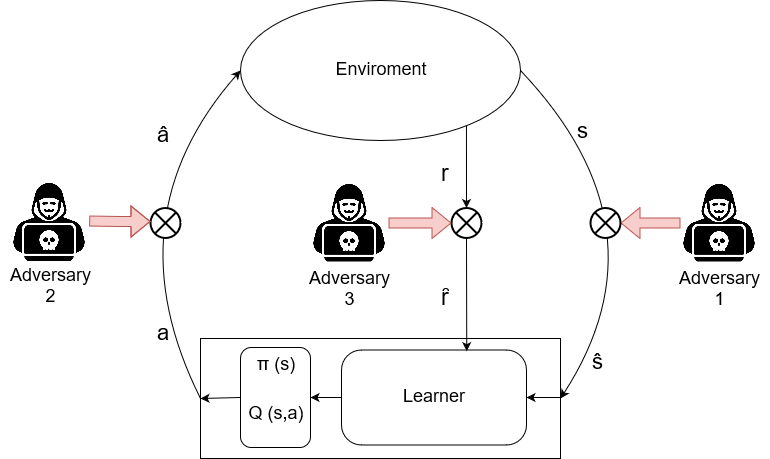}
    \caption{RL proccess during training phase. Each adversary represents a target where the RL training process can be perturbed.}
    \label{fig:RLproccess}
\end{figure}
\begin{enumerate}
    \item \textbf{State perception.} Most works focus on attacking the state perception in training \cite{kos2017delving,pattanaik2018robust} or testing \cite{DBLP:journals/corr/HuangPGDA17,pinto2017robust}. All these works have the common objective of delaying, perturbing or falsifying under malicious attacks the state $s$ perceived by the learning agent. 
    
    \item \textbf{Action selected.} There are also actuator attacks where the objective is to drive the learner agent to undesired states by perturbing at each time step the action $a$ executed by the learning agent~\cite{roy2017reinforcement}.
    
    \item \textbf{Training reward.} Finally, there are reward attacks. These attacks refer to the perturbation of the reward produced by the environment in response to the actions applied by a RL agent~\cite{Everitt2017rc}. 
    
\end{enumerate}

This paper is included in the last category. However, very few works have explicitly studied the modification of the reward function in order to mislead the agent. In one of the first works ~\cite{Everitt2017rc}, the authors propose to modify a set of states where the perceived reward is  different from the real. Then, if the agent reaches these states the enemy performs an attack. Unlike our approach, we establish an attack probability just when the agent achieves a goal state, which significantly reduces the number of attacks. Another possibility is to generate surrogate rewards~\cite{DBLP:journals/corr/abs-1810-01032} from previous noisy rewards to train its agent avoiding the interference of the corruption. Because of this, the learning agent is unable to find the right policy, but also requires a large number of attacks. Instead, modifying the reward with a certain probability following some probability distribution~\cite{romoff2018reward} reduces the attack cost. But in the case where the agent learns without enemy presence, the agent receives wrong reward estimators. Therefore, its performance decreases in a free adversary scenario.  
Another novel approach also modifies rewards driving the learner to the desired goal ~\cite{ma2019policy}. However, unlike our approach, they need more knowledge about the problem, the whole training dataset, and the actions the learner can perform. 
Recently, a framework to perform offline and online attacks on the Stochastic multi-armed bandit problem has been published  ~\cite{liu2019data}. The adversary produces a linear regret on the bandit algorithm with only a logarithmic cost applying their adaptive attack strategy.  As in this approach, our work investigates how to reduce the attacking cost. Therefore, compared to the works mentioned above, our work has a completely different objective: we develop a comprehensive study to characterize how the perturbation probabilities and exploration strategies impact on the convergence of the learning agent in order to find the best configuration from the point of view of the learner but also the adversary. As far as we know, such study has never been carried out before.
    
\end{section}

\begin{section}{Problem Formulation}

We define this task as a stochastic \textit{goal-only-rewards} MDP that is represented by the tuple: (${\cal S}$ , ${\cal A}$, ${\cal G} $, ${\cal P}$, ${r}$, ${\cal \gamma}$, ${s_0}$), where ${\cal S}$ and ${\cal A}$ are state spaces of states $s$ and actions $a$ respectively, and $\cal G \subset S$ is the set of goal states. In the simplest case, \textit{goal-only rewards} are given for states $s \in \cal G$, and zero-reward is given elsewhere. ${\cal P}$ $ : $ ${\cal S} \times {\cal A} \times {\cal S} \rightarrow   {\cal R}$ is the unknown transition probability, $r : {\cal S} \times {\cal A} \rightarrow {\cal R}$ is the reward function, also unknown. $\gamma$ is the discount factor and $s_0$ is the initial state. In our study, we use an on-policy RL algorithm, SARSA. This on-policy algorithm learns the $Q$-value based in the next action performed by the current policy instead of taking the best action as the $Q$-learning algorithm. We update the $Q$ table values with the following function:

\begin{equation}
Q\left(s_{t}, a_{t}\right) \leftarrow(1-\alpha)^{*} Q\left(s_{t}, a_{t}\right)+\alpha\left[r_{t}+\gamma Q\left(s_{t+1}, a_{t+1}\right)\right]
\end{equation}

\noindent where $s_t \in {\cal S}$ represents the current state, $a_t \in {\cal A}$ is the current action performed, $r_t$ is the reward received in this time-step, $s_{t+1} \in {\cal S}$ is the next state and $a_{t+1} \in {\cal A}$ is the action performed in the next state. $\alpha$ is the learning rate.
\subsection{Exploration Strategies and Learner Objective}
In this paper, we study two exploration/exploitation strategies: $\epsilon$-greedy and softmax. $\epsilon$-greedy chooses the best action with a probability $1 - \epsilon$, and a random action with probability $\epsilon$, as shown in Equation~\ref{eq:epsilon}.

\begin{equation}
A \leftarrow\left\{\begin{array}{ll}{\operatorname{argmax}_{a} Q(s, a)} & {\text { with probability } 1-\epsilon} \\ {\text {a random action }} & {\text { with probability } \epsilon}\end{array}\right.
\label{eq:epsilon}
\end{equation}

Instead, softmax assigns a probability of choosing each action as described in Equation~\ref{eq:softmax}. 

\begin{equation}
P(a) = \frac{e^{Q_{t}(a) / \tau}}{\sum_{b=1}^{n} e^{Q_{t}(b) / \tau}}
\label{eq:softmax}
\end{equation}

In Equation~\ref{eq:softmax}, $\tau$ is the temperature parameter. If the value of this parameter is close to 0, the agent will behave totally greedy. On the contrary, if this value is greater, all the actions will be equally probable of performing it.

The goal is to maximize the expected discounted reward per episode as described in Equation~\ref{eq:metric}:

\begin{equation}
\label{eq:metric}
\max_{\pi} (\sum_{t=0}^{N} \gamma^{t} r^{t}) 
\end{equation}

\noindent where $\gamma$ is the discount factor, and $N$ is the maximum number of steps allowed in an episode. It is important to note that this discounted reward per episode will be zero if a goal state is not reached.

\subsection{Threat Model}

We assume that the adversary does not have information about the learner (black-box attack). The adversary can only create bounded perturbations to the true reward signal when the learner reaches a goal state, i.e.,  $\forall s \in \cal G$, $r \neq 0$. In this case, the action of the adversary is limited to changing the sign of the true reward as shown in Equation~\ref{eq:inverseReward}:

\begin{equation}
\label{eq:inverseReward}
    \hat{r}= - r
\end{equation}

\noindent where $r$ is the true reward perceived from the environment, and $\hat{r}$ is the corrupt reward.

Algorithm~\ref{algorithm} describes how the adversary operates during the learning process. The adversary receives as input parameter an attack probability, $p$. Then, it is continuously sniffing the reward channel between the environment and the learning agent as described in Figure~\ref{fig:RLproccess} (line 2 of Algorithm~\ref{algorithm}). If the perceived reward corresponds to a goal state, i.e., $r \neq 0$ (line 3), then the adversary generates a random number $\varphi \in [0,1]$ (line 5). If $\varphi$ is less or equal to $p$, the adversary injects a corrupt reward in the reward channel (lines 6-7). Therefore, at each time $t$ the learning agent reaches a goal state, instead of observing $r$ directly, the agent only observes with probability $p$ a falsified reward $\hat{r}$, ignoring the presence of the adversary.

\begin{algorithm}
\caption{Enemy Injection Algorithm}
\label{algorithm}
\begin{algorithmic}[1]
 \REQUIRE $p$ 
 \REPEAT
 \STATE $r \leftarrow$ \textit{SniffingRewardChannel() }
 \IF{$r \neq 0$} 
 \STATE $\varphi \leftarrow rand() $ 
 \IF{$\varphi \leq p$}
 \STATE $\hat{r} \leftarrow - r$
 \STATE $inject(\hat{r})$
 \ENDIF
 \ENDIF
  \UNTIL stop criterion becomes true
 \RETURN 
\end{algorithmic}
\end{algorithm}

In the experiments, we will evaluate the performance of the adversary according to two criteria. First, the performance of the learning agent according to Equation~\ref{eq:metric}, and then, we count the number of attacks performed. Therefore, the worse the learning agent performance and the lower the number of attacks, the better adversary behavior.

\end{section}

\begin{section}{Evaluation}

The experiments aim to study the behavior of a learning agent and an adversary in a \textit{goal-only reward} MDP. The following sections describe the evaluation environment, the evaluation settings, and the results achieved.

\subsection{Evaluation Environment}

Figure \ref{fig:map} shows an example random 20x20 grid used to perform the experiments. We run a random scenario generator to create different learning episodes.  The green cell denotes the initial state where the learning agent is starting, the red cells denote the goals  it has to reach, and the brown cells denote obstacles. The agent has the actions $\cal{A}=\{\uparrow, \downarrow, \leftarrow, \rightarrow\}$. An action $a \in \cal{A}$ takes the agent in the denoted direction if possible. With a probability of 0.1, the agent is not transported to the desired direction but to one of the three remaining directions. In total, we generate 50 different grids placing the goal states at a similar distance from the starting point. We put these goals between 20 and 25 steps from the initial state.
The results show the average of all the scenarios.

\begin{figure}[ht!]
    \centering
    \includegraphics[scale=0.4]{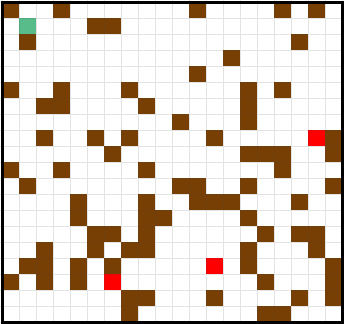}
    \caption{Random map example}
    \label{fig:map}
\end{figure}

The learning agent starts in the initial position and it can move until one of the goal states or until a maximum number number of steps per episode is reached. We establish this maximum number of steps to 500, a enough number of steps to allow the agent to explore and to reach one goal considering the size of the grid. We establish a different reward for each goal $r = \{0.25,\; 0.5\; or \; 1\}$, and $r=0$ otherwise.

\subsection{Evaluation Settings}

We have performed two sets of experiments. In our first experiments, both exploration strategies fix the probability of exploring and exploiting. For the second set of experiments, we dynamically adjust the exploration probability of $\epsilon$-greedy instead of fixing it. In particular, our set of experiments are based on: 
 
\begin{enumerate}

\item {\textbf{Using Fixed Exploration Probabilities.}} In this first set, for each $\epsilon$-greedy experiment, we choose $\epsilon$ values in the range $\epsilon=[0.0,1.0]$, in steps of 0.1.  In respect to softmax, we select the same range values for $\tau$ as $\epsilon$ but ignoring the 0.0 value because it causes a mathematical error. Instead of this value, we choose a value close to zero in order to replicate greedy behavior. In case of tie, we randomly select an action between all the tie actions. Additionally, in this first set of experiments, we consider a learning scenario where the adversary lets the agent learn a policy without perturbing any corrupted rewards and then it starts to craft attacks. In this case, the adversary starts to inject attacks from episode 1.000 to ensure that the learner has learned a policy at this episode.

\item \textbf{{Using Dynamic Exploration Probabilities.}} In this set of experiments, we dynamically adjust the exploration probability, $\epsilon$, of $\epsilon$-greedy instead of fixing it. The training starts with total exploration and ends with total exploitation. We use Equation~\ref{eq:rate} to update the exploration probability at each episode of the learning process. This $\epsilon$ value decreases as a linear function. 

\begin{equation}
    \epsilon_{t} =  1 - \left( \dfrac{e_{t}}{\cal{L}} \right)
    \label{eq:rate}
\end{equation}
where we obtain the dynamic exploration rate dividing the number of the current episode, $e_{t}$, by the number of total episodes of the training phase, $\cal{L}$. In the experiments, we set $\cal{L}$ to 5.000 episodes.
\end{enumerate}

\begin{figure*}[htbp!]
    \centering
    \includegraphics[width=0.9\textwidth]{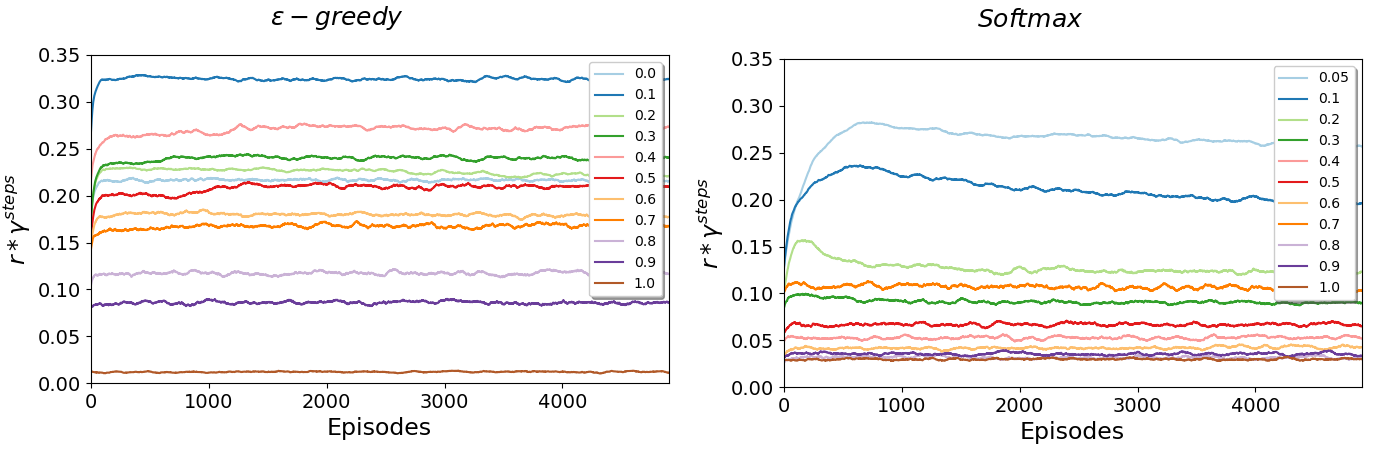}
    \caption{Learning agent performance in an adversary free scenario.}
    \label{fig:baseBehave}
\end{figure*}

The adversary's goal is to produce the maximum deterioration in the learner policy performing the minimum number of attacks, so in the next section we study the results according to two criteria: the performance of the learning agent according to the Equation~\ref{eq:metric}, and the number of attacks injected. Finally, for all experiments $\alpha$ is fixed to 0.125 and $\gamma$ to 0.95, which are classical values in these kind of environments, so the results would be generalizeble to other settings.

\begin{figure*}[hbp!]
    \centering
    \includegraphics[width=0.9\textwidth]{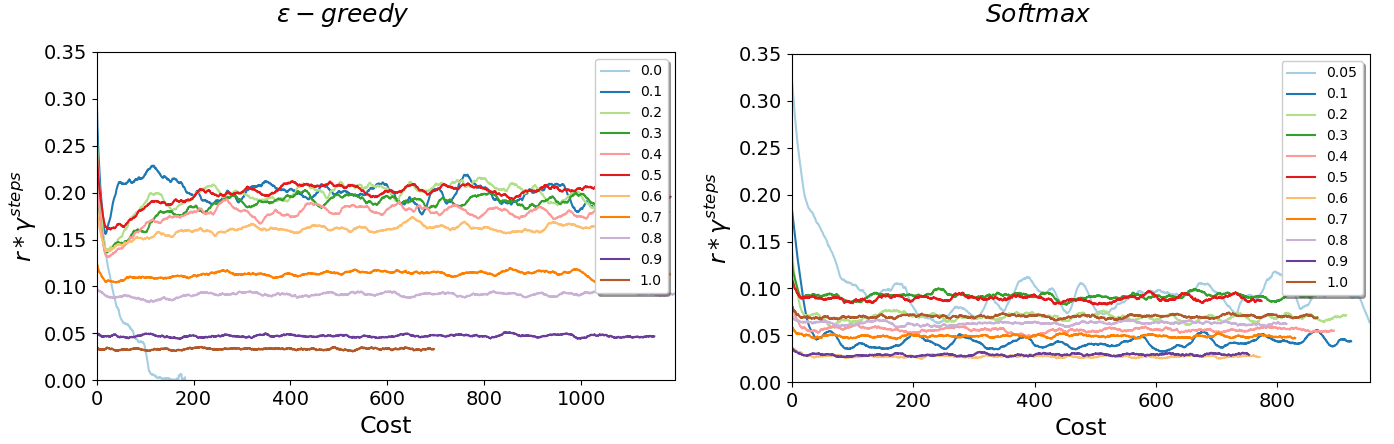}
    \caption{Performance against cost. Configurations under an attack probability $p=0.3$ since episode 1000.}
    \label{fig:cost1}
\end{figure*}

\end{section}

\begin{section}{Results}

\label{sec:results}
Results in this section show the average of ten runs for each configuration and also, we use a slide window of a hundred episodes to reduce the variations between episodes. Following, we describe the results of each set of experiments.

\subsection{Learning without attacks}

First of all, in the first set of experiments, we show the performance of the learner agent without an enemy presence in Figure~\ref{fig:baseBehave}. 
Both plots show different learning curves each one corresponding to a different exploration rate. All configurations reach the goal before a thousand episodes. However, softmax takes more episodes for reaching their maximum values than $\epsilon$-greedy.

\begin{subsection}{Using Fixed Exploration Probabilities}

In Figure~\ref{fig:baseBehave}, both exploration strategies show that the agent reaches the goal in fewer steps using low values of exploration. 
A higher rate of exploration performs more exploratory than exploitative actions and, although the agent still reaches the goal, the path is longer. On the contrary, low exploration rates lead the agent to perform more exploitative than exploratory actions, which cause to fall in local minimum from which the agent is not able to escape. Finally, Figure~\ref{fig:baseBehave} 
also shows that the agent behaves almost randomly when $\epsilon$ or the temperature parameter are larger than 0.7.

Once we analyzed the performance using a \textit{free-of-attacks} scenario, we study how perturbations affect the system. In this set of experiments, the enemy injects adversarial rewards since episode 1000. At this time, a policy to reach one of the goals is learned by both exploration strategies.

In the learning processes of Figure~\ref{fig:attack1}, we fix the attack probability $p$ to $p=0.3$. In the case of $\epsilon$-greedy, Figure~\ref{fig:attack1} shows this attack probability is enough to destroy a total exploitation strategy policy. We observe that all the exploration rates obtain their worst results a few episodes after the emergence of corrupted rewards. After that, the learning processes associated to low exploration rates are able to improve their behaviors but without achieving the initial performances.

A similar analysis can be made with the softmax strategy at the right in Figure \ref{fig:attack1}. Few episodes later the irruption of corrupted rewards the performance of all configurations drops drastically. However, interestingly and in contrast to $\epsilon$-greedy, these agents are not able to slightly recover their performance. Figure~\ref{fig:attack1} demonstrates that low exploration rates are more resilient to small probabilities of attack. The reason is that these configurations allow the agent to reach a goal more frequently and, hence, it still receives more positives than negatives rewards, but it is unable to repair the policy.


Nevertheless, both strategies significantly decrease their performance under corrupted rewards, even with a low attack rate. When the adversary increases this rate, the agent is completely lost with attack probabilities higher that than $p=0.4$. In these cases, all the learned policies are wrong. Agents can occasionally reach the goal cell using exploration rates higher than $0.7$, but it is not enough to repair their policies.


One of the main objective of the adversary is to avoid being detected.  Its objective is to reduce the discounted reward of the learner and also the number of attacks. Each attack has a cost associated. In this problem, we measure the cost as the number of attacks performed by the adversary. In Figure \ref{fig:cost1}, we combine the performance of the agent learner in the vertical axis and the cost in the x-axis. In this way, Figure \ref{fig:cost1} shows the two metrics the adversary wants to optimize. The lower-left the better for the adversary and the other way around for the learner. We show an interesting pattern in the first attacks which are the most harmful to deteriorate the policy.
We confirm that low exploration rates need more attacks to reduce their performances and, also, that few attacks affect considerably the highest exploration rates.

When we increase the attack probability $p$, the performance of all configurations decrease. We show in Figure \ref{fig:cost2} the minimum attack probability which is able to worsen any behavior. Also, we analyze the number of attacks needed to corrupt their policies. However, exploration strategies with exploration rates higher than $0.6$ reach one of the goals randomly. For this reason, we only compare configurations which use the learned policy. Comparing the Figures \ref{fig:cost1} and \ref{fig:cost2}, we observe that the higher the attack rate, the fewer number of attacks needed to destroy the policy. In the Figure \ref{fig:cost1}, the adversary performs more than $1000$ attacks to destroy the policies with the highest exploration rates. On the other plot, the adversary needs only less than 400 attacks to break the configurations which follow a policy, this means that the second attack strategy reduces the number of attacks more than 50\%.
In addition, attacking at each learning episode obtains the lowest cost for all configurations. Here, the adversary is able to destroy all the policies using only half as much as the setting used in the Figure \ref{fig:cost2}.

Due to exploration strategies, softmax keeps reaching at least one of the problem goals.   This explains why softmax performs a higher number of attacks.

\begin{figure*}[htpb!]
  \centering
  \includegraphics[width=0.9\textwidth]{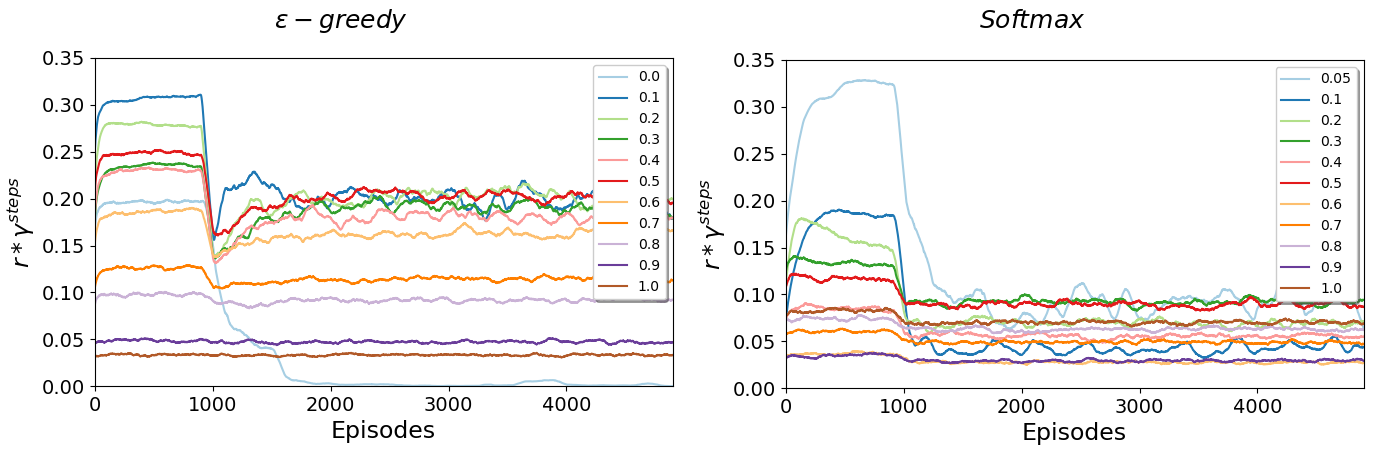}
  \caption{Performance of $\epsilon$-greedy and softmax under an attack probability $p=0.3$.}
  \label{fig:attack1}
\end{figure*}

\begin{figure*}[htpb!]
    \centering
    \includegraphics[width=0.9\textwidth]{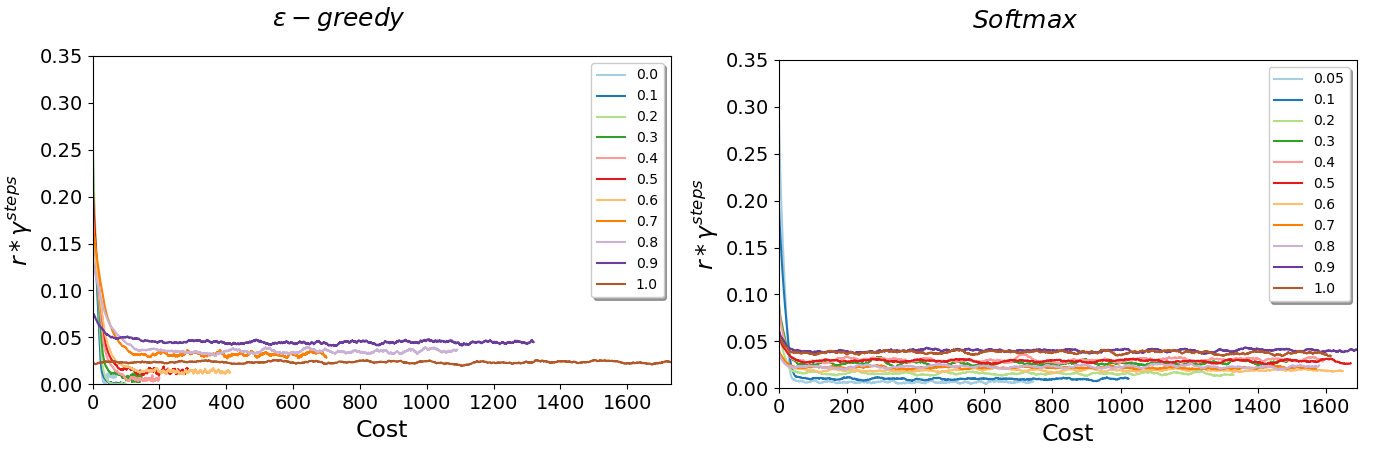}
    \caption{Performance against cost. Configurations under an attack probability $p=0.7$.}
    \label{fig:cost2}
\end{figure*}

The cause of this reduction is that report negative rewards frequently produces changes in the Q-table. The update function changes the value of the optimal actions for one negative and the agent selects actions that keep him away from the goal. For this reason, we consider consecutive attacks as the main factor to create successfully attack strategies.

In the light of the findings, we can conclude that softmax needs more episodes to learn and it suffers from adversarial perturbations further than $\epsilon$-greedy.
In addition, it is required an exhaustive tuning of  its temperature parameter in order to reach an appropiatte performance.

\end{subsection}

\begin{subsection}{Using Dynamic Exploration Probabilities}

In the second set of experiments, we use a dynamic probability of exploration, starting with the maximum exploration rate (i.e., $\epsilon=1$) and ending with total exploitation (i.e., $\epsilon=0$). 
In this way, the agent totally explores until it learns a policy and, in future episodes, starts to follow its policy.

\begin{figure*}[htpb!]
    \centering
    \includegraphics[width=0.9\textwidth]{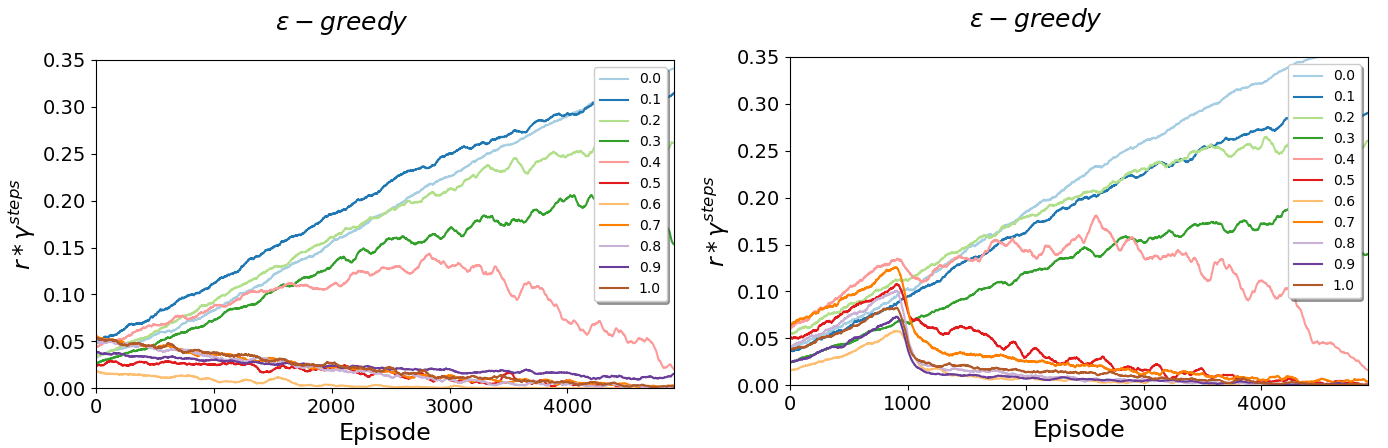}
    \caption{$\epsilon$-greedy using dynamic exploration. At the left, performing attacks from the beginning. And at the right, injecting attacks after episode 1000.}
    \label{fig:dynamic}
\end{figure*}

In Figure \ref{fig:dynamic}, we show the performance during the training updating the exploration rate every episode. In this set of experiments, unlike the former set, each line uses a different attack probability $p$.  We increase this probability from $p=0.0$ to $p=1.0$ in steps of $0.1$. The adversary return corrupted rewards from the beginning of the task. We use the configuration without adversary presence as reference to compare with the rest of experiments. In contrast to the first set of experiments, the agent takes more time  to achieve its maximum performance. In particular, the agent obtains the best result at the end of the training. Focusing on the behavior of the corrupted rewards, we observe that the learner is able to reach the goal in few steps more than using the optimal policy with attack probabilities lower than $p=0.3$.  Using attack probabilities of $p=0.3$ or $p=0.4$ produces that the policies get worsen when the agent reduces significantly its exploration rate. In addition, attack probabilities higher of  $p=0.5$ or higher produce that the agent gets lost in all episodes.

Figure \ref{fig:dynamic} demonstrate that the dynamic configuration is more robust to corrupted rewards than the fixed. In the fixed exploration experiments, we visualize huge variations between episodes when the agent is under enemy perturbations. On the contrary, the configuration with our adaptive exploration strategy achieves an almost identical behavior when the attack rate is low. Additionally, in terms of cost, the adversary needs to increase significantly the number of perturbations. This occurs because the attacked agent is still reaching the goal and the adversary continues crafting corrupted rewards. In this way, the adversary needs a greater number of attacks to destroy the learned policy and the agent may detect the insider.

\end{subsection}

\end{section}

\begin{section}{Conclusion}


In this paper, we execute an exhaustive evaluation to test the robustness of different exploration/exploitation strategies. We use two set of experiments: fixed and dynamic exploration. The former modifies the probability of exploration and we compare these configurations under different adversary attack rates. In the latter, we use a dynamic exploration strategy to analyze the effects under different enemy strategies. Both set of experiments show that the highest probabilities of exploration are not able to reach the goal under the influence of an adversary, even with a low attack probability. Also, increasing the attack probability over the 40\%, the adversary destroys all the policies in a few steps.  As a result, we demonstrate that the first attacks after a learned policy and the consecutive adversary injections are the most harmful to deteriorate the agent behavior.


From the adversary point of view, it aims to avoid being detected, so we propose a method to inject adversarial rewards reducing the number of attacks and the hardness of craft. In terms of cost, the adversary obtains its best performance attacking in each episode. This strategy produces policy failures performing less than 400 corrupted rewards in all the configurations. To decrease the number of attacks, we propose to attack only when the learner reaches the goal state. Also, we modify the attack probability to identify the best configuration for the adversary. Attacking  at  each learning episode is the cheapest adversary strategy. 

The experiments presented in this paper demonstrate some characteristics about exploration strategies in the presence of Adversarial perturbations.

\begin{enumerate}
    \item $\epsilon$-greedy is more robust than softmax. The probability of performing actions that differ from the policy learned in addition to the complexity of finding the right temperature parameters cause softmax obtains worse results than $\epsilon$-greedy. 
    \item Low exploration ratios are more resilient to attacks than larger values. Larger values of exploration produce a significant number of random actions aiming to reach any goal, regardless the number of steps or even the reward the agent receives.
    \item Dynamic exploration strategy is more resistant to perturbations. The dynamic exploration agent is able to resist  perturbation rates below 0.3 even from the beginning of the training and it obtains a performance similar to the free adversary scenario. We consider that this strategy is an excellent alternative to train an agent when there are chances of obtaining disturbances in the reward function.
    \item As opposed to what one might think, performing attacks every episode becomes the cheapest option. Performing consecutive attacks are more harmful because the agent can not adjust their Q table values to correct these perturbations in future episodes.
\end{enumerate}




In future work, we would generalize our approach using other domains such as the Atari games used in Deep RL. Also, we would analyze the behavior of other agent configurations. Testing more elaborate attacks strategies such as introducing intermediate rewards or developing smart adversary behaviors are some examples of an exciting topic for future research.

\end{section}

\begin{section}{Acknowledgment}

This research is granted by Repsol, the Spanish Goverment (Ministerio de Economia y Empresa) and FEDER, UE funds under project TIN2017-88476-C2-2-R.
This research was funded in part by JPMorgan Chase \& Co. Any views or opinions expressed herein are solely those of the authors listed, and may differ from the views and opinions expressed by JPMorgan Chase \& Co. or its affiliates. This material is not a product of the Research Department of J.P. Morgan Securities LLC. This material should not be construed as an individual recommendation for any particular client and is not intended as a recommendation of particular securities, financial instruments or strategies for a particular client. This material does not constitute a solicitation or offer in any jurisdiction.

\end{section}

%
%
%
\bibliographystyle{ecai}
\bibliography{references}

\begin{thebibliography}{10}

\bibitem{behzadan2017vulnerability}
Vahid Behzadan and Arslan Munir, `Vulnerability of deep reinforcement learning
  to policy induction attacks', in {\em International Conference on Machine
  Learning and Data Mining in Pattern Recognition}, pp. 262--275. Springer,
  (2017).

\bibitem{tong2019}
Tong Chen, Jiqiang Liu, Yingxiao Xiang, Wenjia Niu, Endong Tong, and Zhen Han,
  `Adversarial attack and defense in reinforcement learning-from ai security
  view', {\em Cybersecurity}, {\bf 2}, (12 2019).

\bibitem{Everitt2017rc}
Tom Everitt, Victoria Krakovna, Laurent Orseau, Marcus Hutter, and Shane Legg,
  `Reinforcement learning with a corrupted reward signal', in {\em Proceedings
  of the Twenty-Sixth International Joint Conference on Artificial
  Intelligence, {IJCAI} 2017, Melbourne, Australia, August 19-26, 2017}, pp.
  4705--4713, (2017).

\bibitem{ferdowsi2018robust}
Aidin Ferdowsi, Ursula Challita, Walid Saad, and Narayan~B Mandayam, `Robust
  deep reinforcement learning for security and safety in autonomous vehicle
  systems', in {\em 2018 21st International Conference on Intelligent
  Transportation Systems (ITSC)}, pp. 307--312. IEEE, (2018).

\bibitem{DBLP:journals/corr/HuangPGDA17}
Sandy~H. Huang, Nicolas Papernot, Ian~J. Goodfellow, Yan Duan, and Pieter
  Abbeel, `Adversarial attacks on neural network policies', {\em CoRR}, {\bf
  abs/1702.02284}, (2017).

\bibitem{kos2017delving}
Jernej Kos and Dawn Song, `Delving into adversarial attacks on deep policies',
  {\em arXiv preprint arXiv:1705.06452}, (2017).

\bibitem{kuderer2015learning}
Markus Kuderer, Shilpa Gulati, and Wolfram Burgard, `Learning driving styles
  for autonomous vehicles from demonstration', in {\em 2015 IEEE International
  Conference on Robotics and Automation (ICRA)}, pp. 2641--2646. IEEE, (2015).

\bibitem{DBLP:journals/jmlr/LevineFDA16}
Sergey Levine, Chelsea Finn, Trevor Darrell, and Pieter Abbeel, `End-to-end
  training of deep visuomotor policies', {\em J. Mach. Learn. Res.}, {\bf 17},
  39:1--39:40, (2016).

\bibitem{liu2019data}
Fang Liu and Ness Shroff, `Data poisoning attacks on stochastic bandits', in
  {\em International Conference on Machine Learning}, pp. 4042--4050, (2019).

\bibitem{ma2019policy}
Yuzhe Ma, Xuezhou Zhang, Wen Sun, and Xiaojin Zhu, `Policy poisoning in batch
  reinforcement learning and control', in {\em Advances in Neural Information
  Processing Systems}, (2019).

\bibitem{mnih2013playing}
Volodymyr Mnih, Koray Kavukcuoglu, David Silver, Alex Graves, Ioannis
  Antonoglou, Daan Wierstra, and Martin Riedmiller, `Playing atari with deep
  reinforcement learning', {\em arXiv preprint arXiv:1312.5602}, (2013).

\bibitem{morere18a}
Philippe Morere and Fabio Ramos, `Bayesian rl for goal-only rewards', in {\em
  Proceedings of The 2nd Conference on Robot Learning}, eds., Aude Billard,
  Anca Dragan, Jan Peters, and Jun Morimoto, volume~87 of {\em Proceedings of
  Machine Learning Research}, pp. 386--398. PMLR, (29--31 Oct 2018).

\bibitem{pattanaik2018robust}
Anay Pattanaik, Zhenyi Tang, Shuijing Liu, Gautham Bommannan, and Girish
  Chowdhary, `Robust deep reinforcement learning with adversarial attacks', in
  {\em Proceedings of the 17th International Conference on Autonomous Agents
  and MultiAgent Systems}, pp. 2040--2042. International Foundation for
  Autonomous Agents and Multiagent Systems, (2018).

\bibitem{pinto2017robust}
Lerrel Pinto, James Davidson, Rahul Sukthankar, and Abhinav Gupta, `Robust
  adversarial reinforcement learning', {\em arXiv preprint arXiv:1703.02702},
  (2017).

\bibitem{reinke2017}
Chris Reinke, Eiji Uchibe, and Kenji Doya, `Average reward optimization with
  multiple discounting reinforcement learners', in {\em Neural Information
  Processing}, eds., Derong Liu, Shengli Xie, Yuanqing Li, Dongbin Zhao, and
  El-Sayed~M. El-Alfy, pp. 789--800, Cham, (2017). Springer International
  Publishing.

\bibitem{romoff2018reward}
Joshua Romoff, Alexandre Pich{\'e}, Peter Henderson, Vincent Francois-Lavet,
  and Joelle Pineau, `Reward estimation for variance reduction in deep
  reinforcement learning', {\em arXiv preprint arXiv:1805.03359}, (2018).

\bibitem{roy2017reinforcement}
Aurko Roy, Huan Xu, and Sebastian Pokutta, `Reinforcement learning under model
  mismatch', in {\em Advances in Neural Information Processing Systems}, pp.
  3043--3052, (2017).

\bibitem{DBLP:journals/corr/SzegedyZSBEGF13}
Christian Szegedy, Wojciech Zaremba, Ilya Sutskever, Joan Bruna, Dumitru Erhan,
  Ian Goodfellow, and Rob Fergus, `Intriguing properties of neural networks',
  in {\em International Conference on Learning Representations}, (2014).

\bibitem{alphastarblog2}
Oriol Vinyals, Igor Babuschkin, Junyoung Chung, Michael Mathieu, Max Jaderberg,
  Wojciech~M. Czarnecki, Andrew Dudzik, Aja Huang, Petko Georgiev, Richard
  Powell, Timo Ewalds, Dan Horgan, Manuel Kroiss, Ivo Danihelka, John Agapiou,
  Junhyuk Oh, Valentin Dalibard, David Choi, Laurent Sifre, Yury Sulsky, Sasha
  Vezhnevets, James Molloy, Trevor Cai, David Budden, Tom Paine, Caglar
  Gulcehre, Ziyu Wang, Tobias Pfaff, Toby Pohlen, Yuhuai Wu, Dani Yogatama,
  Julia Cohen, Katrina McKinney, Oliver Smith, Tom Schaul, Timothy Lillicrap,
  Chris Apps, Koray Kavukcuoglu, Demis Hassabis, and David Silver.
\newblock {AlphaStar: Mastering the Real-Time Strategy Game StarCraft II},
  2019.

\bibitem{DBLP:journals/corr/abs-1810-01032}
Jingkang Wang, Yang Liu, and Bo~Li, `Reinforcement learning with perturbed
  rewards', {\em CoRR}, {\bf abs/1810.01032}, (2018).

\end{thebibliography}

\end{document}